\documentclass[journal=jacsat,manuscript=article]{achemso}
\usepackage{graphicx}
\usepackage{subcaption}
\usepackage[version=3]{mhchem} 
\usepackage{siunitx}
\usepackage[dvipsnames]{xcolor}
\usepackage{amssymb}
\usepackage{pifont}
\usepackage{wrapfig}
\usepackage{booktabs}
\usepackage{multirow}
\usepackage{hyperref}
\usepackage{microtype}
\usepackage{tabularray}
\usepackage{geometry}
\usepackage{array}
\usepackage{makecell}

\usepackage{amsmath} 

\author{Saurabh Patil}
\affiliation[meche]
{Department of Mechanical Engineering, Carnegie Mellon University, USA}
\author{Zijie Li}
\affiliation[meche]
{Department of Mechanical Engineering, Carnegie Mellon University, USA}
\author{Amir Barati Farimani}
\email{barati@cmu.edu}
\affiliation[meche]
{Department of Mechanical Engineering, Carnegie Mellon University, USA}
\alsoaffiliation[biomed]
{Department of Biomedical Engineering, Carnegie Mellon University, USA}
\alsoaffiliation[mld]
{Machine Learning Department, Carnegie Mellon University, USA}

\title[An \textsf{achemso} demo]
  {Hyena Neural Operator for Partial Differential Equations}

\abbreviations{IR,NMR,UV}

\begin{document}







\begin{abstract}
Numerically solving partial differential equations typically requires fine discretization to resolve necessary spatiotemporal scales, which can be computationally expensive. Recent advances in deep learning have provided a new approach to solving partial differential equations that involves the use of neural operators. Neural operators are neural network architectures that learn mappings between function spaces and have the capability to solve partial differential equations based on data. This study utilizes a novel neural operator called Hyena, which employs a long convolutional filter that is parameterized by a multilayer perceptron. The Hyena operator is an operation that enjoys sub-quadratic complexity and state space model to parameterize long convolution that enjoys a global receptive field. This mechanism enhances the model's comprehension of the input's context and enables data-dependent weight for different partial differential equations instances.  To measure how effective the layers are in solving partial differential equations, we conduct experiments on Diffusion-Reaction equation and Navier Stokes equation. Our findings indicate Hyena Neural operator can serve as an efficient and accurate model for learning partial differential equations solution operator. The data and code used can be found at: \url{https://github.com/Saupatil07/Hyena-Neural-Operator}
\end{abstract}

\section{Introduction}
Numerical modeling of Partial differential equations (PDEs) plays a crucial role in engineering as they serve as fundamental tools for representing and analyzing various physical phenomena. They find application in diverse areas, such as fluid dynamics, gas dynamics, electrical circuitry, heat transfer, and acoustics, enabling us to model and understand these phenomena effectively. PDEs provide a framework for understanding complex systems by describing the relationships between various quantities that change over time and space. They are widely used in science and engineering to make predictions, optimize designs, and analyze data. Traditional numerical solvers for partial differential equations (PDEs) are often costly because they rely on methods that require a fine discretization of the problem domain. Numerous techniques in deep learning have been proposed to address the computational complexity of numerical solvers and to forecast fluid properties. These approaches include reinforcement learning\cite{foucart2022deep,pmlr-v206-yang23e,meshdqn}, surrogate modeling\cite{dl_rom,hemmasian_surrogate_2023}, generative adversarial networks (GANs)\cite{farimani2017deep,fluid_superresolution,volumetric_gan,Werhahn_2019} and diffusion models.\cite{shu_physics-informed_2023,yang2023denoising,wang2023generative,jadhavstressd} 
Neural operators are designed to operate on function representations and enable the learning of operators directly from data. Compared to traditional solvers, they alleviate the need for fine discretization and can be used to infer the solution of different instances within a family of PDE once trained. 
One of the earliest neural operators proposed was the DeepONet \cite{lu2019deeponet}. It consists of a branch network responsible for processing the input functions and learning the action of the operator, along with a trunk network that learns the function bases for the solution function space. \citet{wang2021learning} further improved the performance of DeepONets by introducing an improved architecture and training methods. MIONet \cite{jin2022mionet} extends DeepONet to problems involving multiple input functions. In addition to DeepONet, another group of methods \citep{kovachki2023neural, li2020neural} leverage a learnable kernel integral to approximate the target operator. A notable instance is Fourier Neural Operator \citep{li2020fourier}(FNO), which utilizes the Fourier transform to learn the convolution kernel integral in the frequency domain. The Fourier neural operator has been further adapted to various forms as shown in (\citet{tran2023factorized,guibas2022efficient,li2023physicsinformed}). Other than the Fourier domain, the wavelet domain has also been explored in (\citet{tripura2022wavelet,gupta2021multiwavelet}). \citet{cao2021choose} draws the connection between a softmax-free attention and two different types of integral and proposes a attention-based operator learning framework. \citet{li2022transformer} further expands the work on attention by proposing to propagate to the solution in latent space with cross-attention mechanism and relative positional encoding\cite{su2022roformer}.

 Various previous works \cite{li2020fourier, cao2021choose, gupta2021multiwavelet} have shown that the capability of capturing global interaction is crucial to the prediction accuracy. Non-local learnable modules such as spectral convolution \citep{li2020fourier}, attention \citep{cao2021choose} or dilated convolution \citep{stachenfeld2022learned} are better at learning complex time-evolving dynamics where other local learnable modules like residual neural network (ResNet)\citep{He_2016_CVPR} often fails to model. State space models (SSMs) are a type of recurrent model that can be viewed as long-context convolution. It effectively extends the receptive field to the whole input sequence and has the potential to learn and model complex non-local interaction that lies in the PDE data. 
 
 The state space models are represented by the following equations: 
\begin{equation}
    x(t+1) =Ax(t)+Bu(t), \quad
    y(t)=Cx(t)+Du(t),
\end{equation}
where the input $u(t)$ represents a one-dimensional signal, while the state $x(t)$ represents an N-dimensional hidden representation that follows a linear ordinary differential equation (ODE). The output $y(t)$ is a straightforward one-dimensional projection of the state. $A, B, C, D$ are learned projections.
State space models \cite{chen1984linear} serve as a foundational framework widely employed in scientific and engineering fields like control theory. Earlier examples of SSM layers in deep learning model includes Structured State Space(S4)\citep{gu2022efficiently}, its variants \citep{gu2022parameterization, gu2022train} and Gated State Space (GSS) \citep{mehta2022long}.   A later work Hungry Hungry Hippo (H3)\cite{dao2022hungry} was proposed to address the limitations of prior SSM layers, specifically targeting two key drawbacks: their incapability to recall previous tokens in the sequence and their expensive computational cost. H3 solves the associative recall by including an additional gate and a short convolution obtained via a shift SSM. It also proposes FlashConv, a fast and efficient algorithm for training and inferring SSMs. It works by using a fused block FFT algorithm to compute the convolutions in the SSM, which significantly reduces the training and inference time. 
Recent work Hyena \cite{poli2023hyena} further extends H3 and incorporates implicit filter parametrization, advancing the accuracy and efficiency of SSM-based model, which have achieved state-of-the-art performance across benchmarks like LRA\cite{tay2020long}.

This work presents a novel deep-learning architecture for learning PDE solutions called Hyena Neural Operator (HNO), which utilizes long convolutions and element-wise multiplicative gating mechanism. Hyena Neural Operator(HNO) employs an Encoder-Decoder architecture with a latent-marching strategy\citep{li2022transformer}. We demonstrate that HNO has competitive performance against Fourier Neural Operator on various numerical benchmarks. 
\begin{figure}[t]
    \begin{center}
    \includegraphics[width=0.9\linewidth]{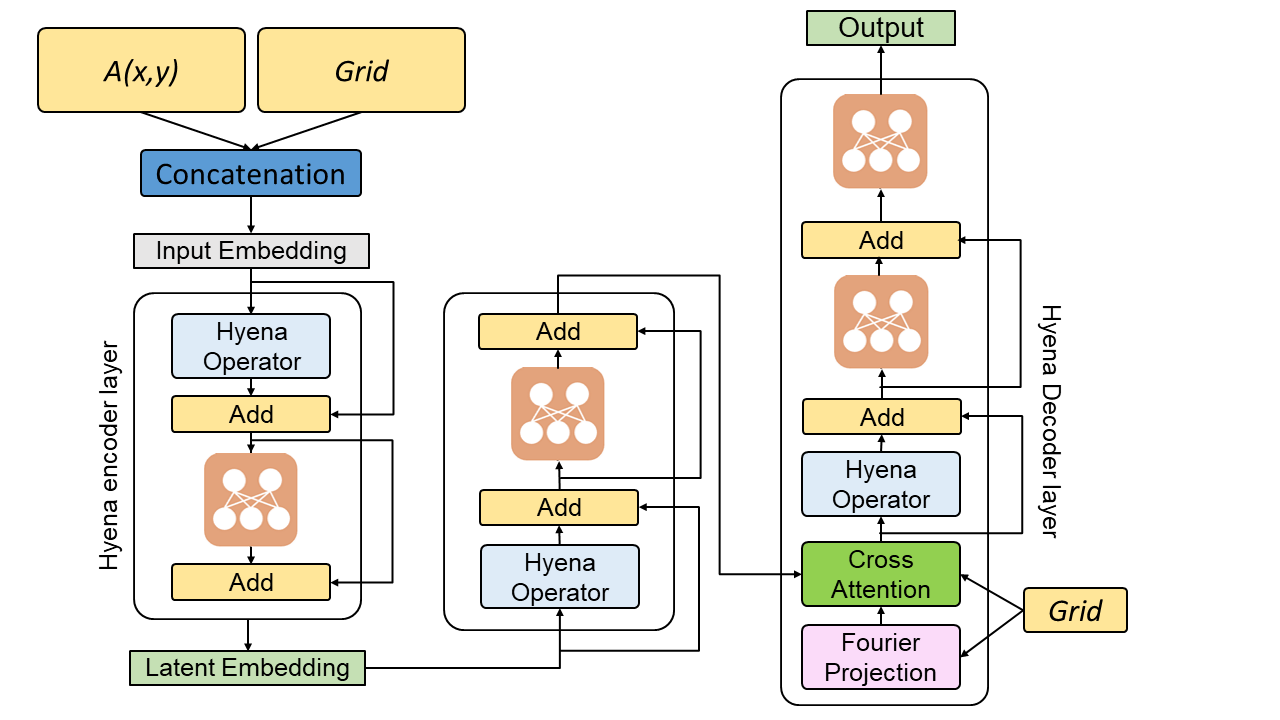}
    \vspace{-2mm}
    \caption{\textbf{Hyena Neural Operator architecture}. Given the initial observation and the grid, the encoder layer encodes it to a latent embedding, which is an input to the latent Hyena layers. The latent output from Hyena layers, Fourier projection, and the grid is given as input to the cross-attention module. The resultant values are once again passed through Hyena layers and the output solution is obtained following an MLP layer.} 
    \label{fig: model architecture}
    \vspace{-4mm}
    \end{center}
\end{figure}
\begin{figure}[t]
    \begin{center}
    \includegraphics[width=0.9\linewidth]{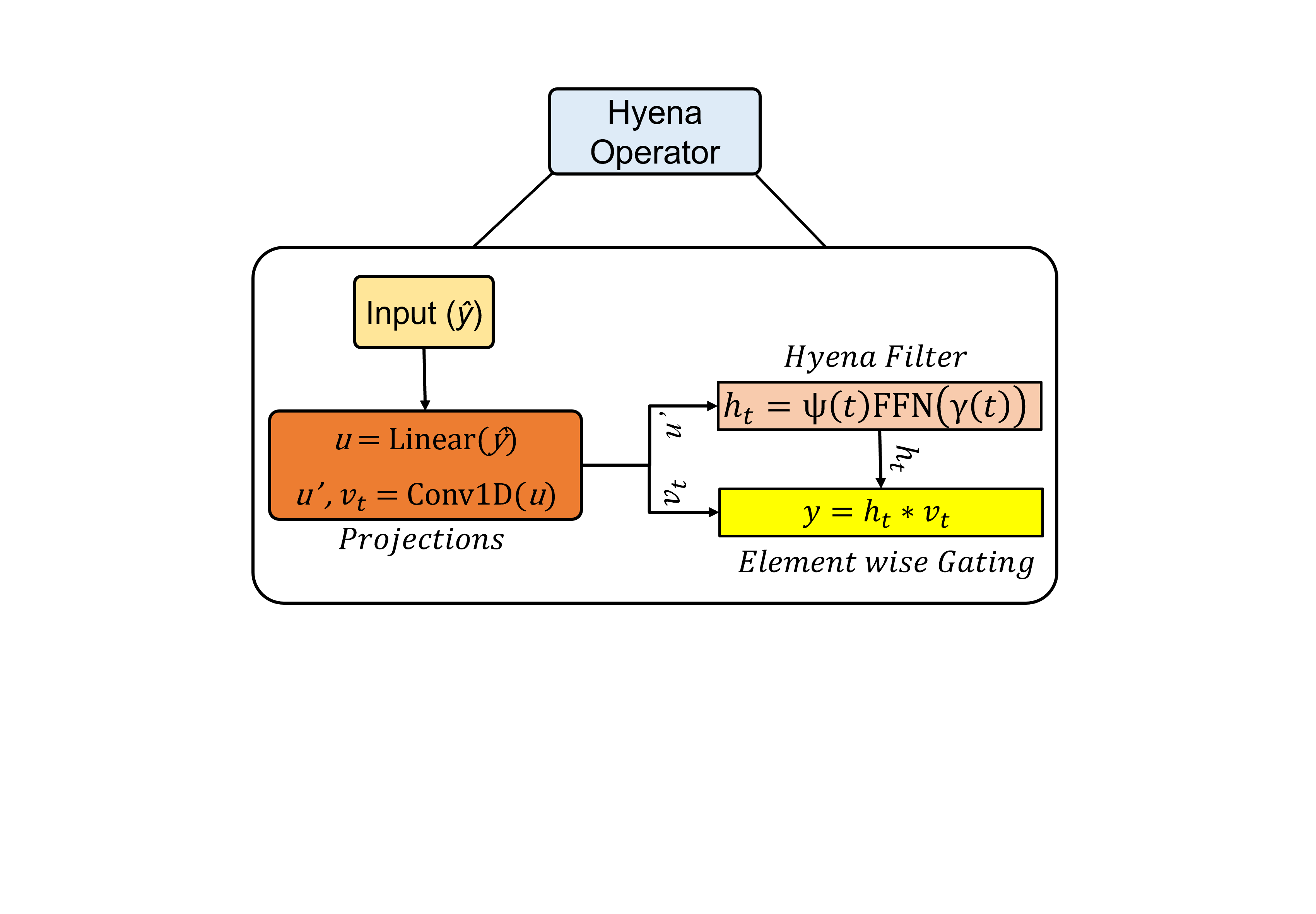}
    \vspace{-36mm}
    \caption{\textbf{Hyena architecture}. The input to the Hyena operator is first projected to a width defined by the order and input dimension. The projections are first passed through a short filter and then to generated filters made on the fly. Inside the Hyena filter, the data is processed in three steps: first the positional encoding, second the implicit filter, and lastly the exponential modulation. 
    } 
    \label{fig: hyena arch}
    \vspace{-4mm}
    \end{center}
\end{figure}
\section{Method}
\subsection{Hyena Neural Operator}
The Hyena operator can be characterized as a repetition of two sub-quadratic operations: an implicit long convolution $h$ (which means that the Hyena filters are implicitly parameterized by the output of a feed-forward network) and a multiplicative component-wise control of the (projected) input. Hyena first computes $N+1$ learnable projections\footnote{In practice it is implemented as a single convolution layer.} of the input: $(v, \xi^1, \cdots, \xi^N)$, which is similar to query/key/value projections in a standard attention mechanism. The next step is to compute the convolution filters, which are implicitly parametrized \citep{mildenhall2021nerf, sitzmann2020implicit, romero2022ckconv} and modulated via a window function. Concretely, the value of the filter $h$ on the $t$-th location is given by:
\begin{equation}
    h_t= \psi(t)\text{FFN}(\gamma(t)),
\end{equation}
where $\psi(\cdot)$ is a window function that decays exponentially with respect to $t$: $\psi(t) = \exp(-\alpha t)$, with $\alpha$ controlling the decaying speed, $\text{FFN}$ denotes the feed-forward network equipped with a sine activation function, 
and $\gamma(\cdot)$ is a positional encoding function: 
\begin{equation}
\gamma(t)=[t, \cos{(2\pi t/L)}, \hdots, \cos{(2\pi Kt/L)}, \sin{(2\pi t/L)}, \hdots, \sin{(2\pi Kt/L)}],
\end{equation}
with $K$ as a hyperparameter, $L$ being the length of the input sequence. The implicit filter decouples the parameter size of the filter and its valid receptive field. The sine activation function together with the positional encoding function allows the filter to learn high-frequency patterns \citep{tancik2020fourier} whereas the exponential decaying function enables the learned filter to focus on the different parts of the input at different steps.

With the computed filter $(h^1, h^2, \cdots, h^N)$ and the projected inputs $(v, \xi^1, \cdots, \xi^N)$, the update rule within a Hyena operator block is defined as follows:
\begin{equation}
    z^{n+1} = \xi^n\odot \mathcal{K}(h^n, z^n), \quad n = 1,....,N, 
    \label{eq: hyena update}
\end{equation}
where $\mathcal{K}$ denotes the convolution operation: $\mathcal{K}(h, u)=h*u=\sum_{n=1}^{L} h_{t-n}u_n$, and $\odot$ denotes element-wise multiplication, $N$ is a hyperparameter. If we view the input sequence as the sampling of a function on the discretization grid $\{x_t\}_{t=1}^N$, then \eqref{eq: hyena update} can be viewed as an approximation to the integral transform:
$z^{n+1}(x_t) = \xi^n(x_t)\int_{\Omega}h^{n}(x_t-y)z^{n}(y)dy$, where the function are iteratively updated by a kernel integral and an instance-based weight value $\xi^n(x_t)$. The spectral convolution layer in FNO can be viewed as a special case of \eqref{eq: hyena update} with filter's value explicitly parameterized and no instance-based weight. 


\paragraph{Encoder} 
The encoder is composed of three main components, an input embedding layer that takes in the input function's sampling and lifts the input features into high-dimensional encodings $\mathbf{u}^{(0)}$, multiple layers of Hyena operator followed by feedforward networks. The output from each Hyena layer is aggregated and then passed on to the projection layer which projects the output from the Hyena layers to latent embedding. The latent embeddings are passed through a series of Hyena layers and the output from the layers is once again aggregated and passed to the decoder. The update protocol inside each Hyena operator block is:
\begin{equation}
    \label{eq:encoding hyena}
    \mathbf{u}^{(l')} = \mathbf{k}^{(l)} + \text{Norm}\left(\text{Hyena}(\mathbf{u}^{(l)})\right), \quad
    \mathbf{u}^{(l+1)} = \text{FFN}(\mathbf{u}^{(l')}),
\end{equation}
where $\text{Hyena}(\cdot)$ denotes the Hyena operator, $\text{Norm}(\cdot)$ denotes the layer normalization layer \citep{ba2016layer}. 
\paragraph{Decoder} 

To generate the solution, the decoder utilizes the input coordinates and the output obtained from the encoder. The first layer is a random Fourier projection layer\citep{tancik2020fourier, random_fourier_feature}. By incorporating random Fourier projection, the inherent spectral bias found in coordinate-based neural networks is alleviated \citep{tancik2020fourier,mildenhall2021nerf}. Following the Fourier projection, the latent encoding $\mathbf{u}^{(L)}$, along with the encoding of positions $\mathbf{p}^{(0)}$ that has been learned, is fed into the cross-attention module inspired by the \citet{li2022transformer}. Finally, the decoder outputs the prediction by taking the result of the cross-attention module, passing it through the Hyena operator, and then applying a feed-forward network. The decoder process can be described as follows:
\begin{equation}
    \mathbf{p}' = \mathbf{p}^{(0)} + \text{Cross-Attn}(\mathbf{p}^{(0)}, \mathbf{u}^{(L)}),
    \label{eq:cross-attn ffn} \quad
    \mathbf{p}'' = \mathbf{p}' + \text{Hyena}(\mathbf{p}'), \quad
    \mathbf{p} = \mathbf{p}'' + \text{FFN}(\mathbf{p}'').
\end{equation}

\paragraph{Training settings}
The overall training framework of this work shares similarities with previous data-driven models focused on operator learning. We used the Adam optimizer\cite{kingma2014adam} and a CosineAnnealing scheduler\cite{loshchilov2017sgdr} with a decay rate of $1e-8$. The dropout rate was set to $0.03$ inside the feedforward layers of the Hyena operator. Unless stated we have trained the models for 500 epochs with an initial learning rate set as \num{1e-4}. We use GELU\cite{hendrycks2020gaussian} activation. To train the model on 2D Navier-Stokes data, we employ a curriculum strategy that involves gradually increasing the prediction time steps following \citet{li2022transformer}. Instead of forecasting all upcoming states until the end of the specified time horizon, we initially limit the duration by a fraction called $\gamma$ (around $\gamma \approx 0.5$) and then gradually grow the time duration as the training progresses. In this approach, the network is trained to predict the states $u_{t_0}, u_{t_1}, \ldots, u_{\gamma T}$. We found that implementing the above strategy worked better than asking the model to predict the whole sequence at once which consequently improves stability and leads to slightly faster convergence.
\begin{figure}[t]
    \centering
    \includegraphics[width=0.6\textwidth]{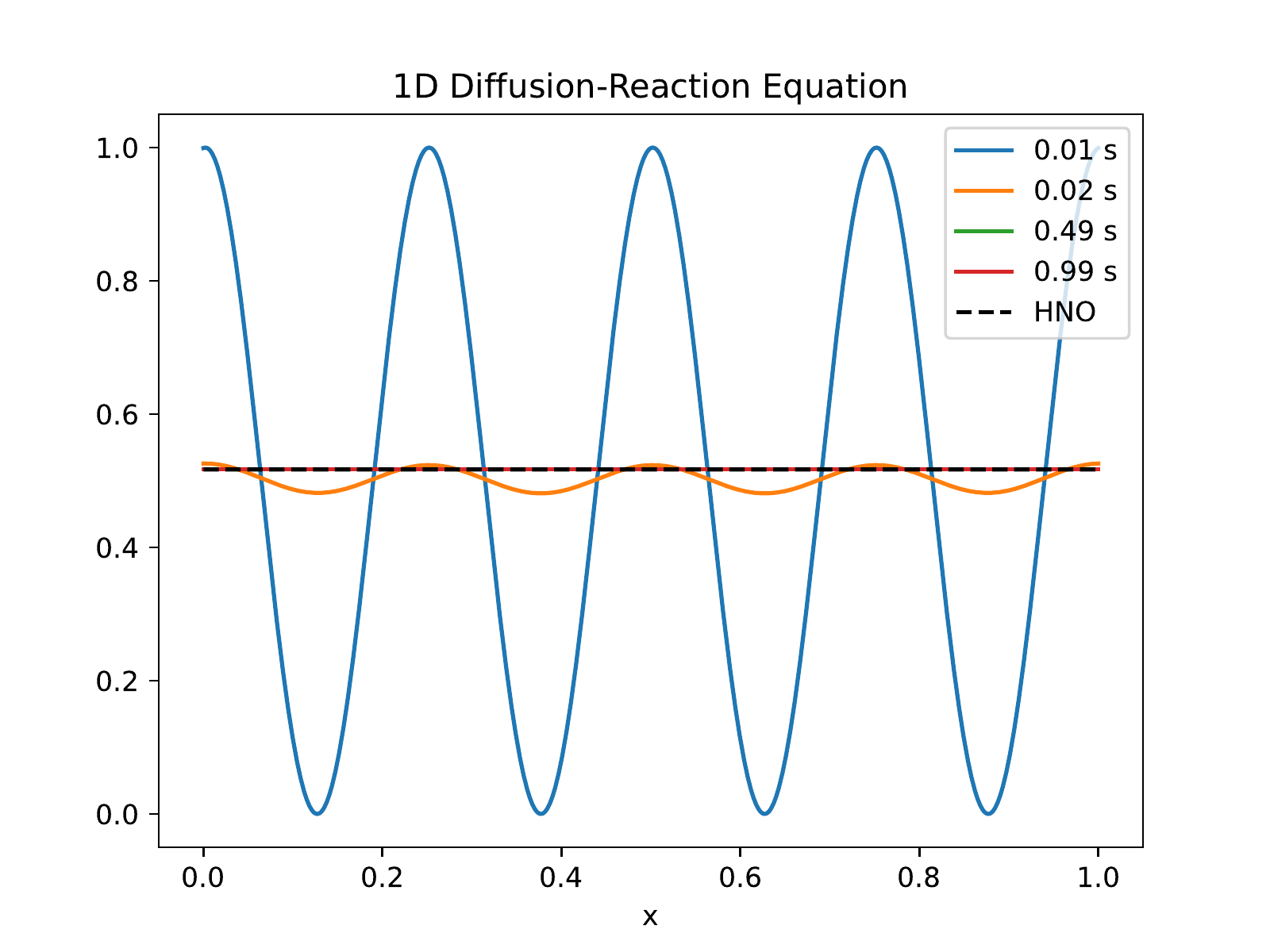}
    \caption{Visualization of time evolution of 1D Diffusion-Reaction equation. Black dotted lines denote the model's output. The blue line denote the initial condition given as input to the models.}
    \label{fig:react_diff}
\end{figure}
\section{Numerical Experiments}
We assess our model's performance using benchmark problems, we consider the 2D Navier-Stokes equation and the 1D Diffusion-Reaction equation. To ensure a comprehensive assessment, we conduct a comparative analysis between the performance of our model and that of the state-of-the-art neural operator, namely the Fourier neural operator. Detailed insights into our model's architecture for different problems are available in Appendix A.

\subsection{1D Diffusion-Reaction}
We used the dataset provided by PDEBench\citep{takamoto2022pdebench} a benchmark for SciML. The data consist of an one-dimensional diffusion-reaction type PDE, that combines a diffusion process and a rapid evolution from a source term \cite{krishnapriyan2021characterizing}. 
The equation is expressed as:
 \begin{align}
     \partial_t u(t,x) & - \nu \partial_{xx} u(t,x) - \rho u (1 - u) = 0,  ~~~ x  \in (0,1), t \in (0,1], \\
     u(0,x) &= u_0(x), ~~~ x \in (0,1).
 \end{align}
We evaluate the performance of Fourier neural operator and Hyena neural operator on different values of $\nu=0.5, 2.0$ at different resolutions. We provide the condition at the initial time step and the model predicts the solution at the final time step. Fig \ref{fig:react_diff} shows the time evolution of the equation. The models have been trained for 200 epochs with a batch size of 20. Table \ref{table: 1d_react} shows that Hyena neural operator consistently performs better than FNO for different values of $\nu$ at varying resolutions. 
\begin{table}[t]
\vspace{-1mm}
\begin{center}
\scalebox{0.95}{
\begin{tabular}{cc!{\vrule width \lightrulewidth}ccc!{\vrule width \lightrulewidth}c} 
\toprule
\multicolumn{2}{c!{\vrule width \lightrulewidth}}{Data settings}          & \multicolumn{4}{c}{Relative $L_2$ norm}                                          \\ 
\midrule
    Case       & Resolution  & FNO & HNO  \\ 
\midrule
$\nu=0.5, \rho=1.0$ & $256$ &   40.16&    \textbf{21.68}\\
 & $512$ &    46.29    &    \textbf{23.89}   \\
 & $1024$   &    42.71    &  \textbf{22.52} \\
 $\nu=2.0, \rho=1.0$ & $256$ &   41.33     &    \textbf{36.32} \\

 & $512$ &    39.20    &    \textbf{36.91}  \\
 & $1024$   &    41.98    &  \textbf{22.78} \\
\bottomrule
\end{tabular}}
\caption{ Relative $L_{2}$ norm({\small$\times 10^{-4}$}) for 1D Diffusion-Reaction. \textbf{Bold} indicates best performance. }\label{table: 1d_react}
\vspace{+2mm}
\end{center}
\vspace{-2mm}
\end{table}

\begin{figure}[t]
    
    \begin{subfigure}{\textwidth}
    \centering
    \vspace{-3mm}
    \includegraphics[width=\linewidth]{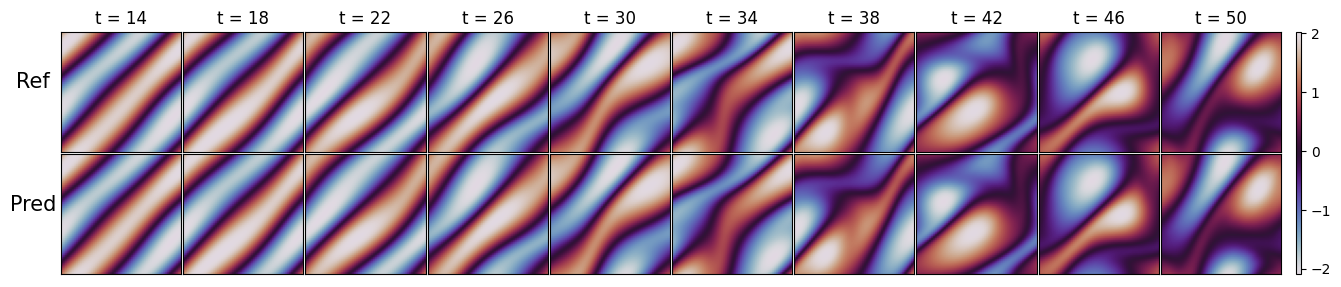}
    \captionsetup{width=0.85\linewidth}
    \vspace{-1.5mm}
    \caption{HNO's prediction for $\nu=1e-3$}
    \end{subfigure}
    \begin{subfigure}{\textwidth}
    \centering
    \includegraphics[width=\linewidth]{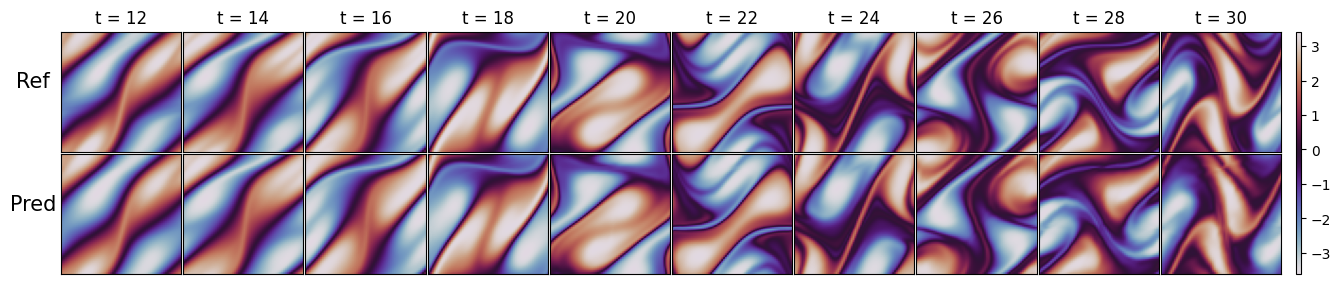}
    \vspace{-1.5mm}
    \caption{HNO's prediction for  $\nu=1e-4$}
    \end{subfigure}
    \begin{subfigure}{\textwidth}
    \centering
    \includegraphics[width=\linewidth]{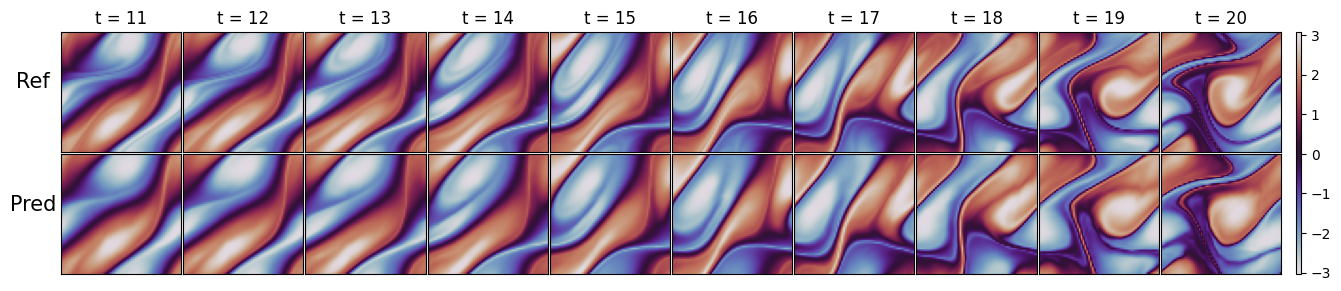}
    \vspace{-1.5mm}
    \caption{HNO's prediction for  $\nu=1e-5$}
    \end{subfigure}
    \vspace{-1.0mm}
    \caption{HNO's prediction on Navier stokes equation.}
    \label{fig: ns_result}
\end{figure}

\subsection{Navier-Stokes Equation}
 
The Navier-Stokes equations are one of the most important equations in physics. They are a fundamental description of the motion of fluids. It is a complex and nonlinear equation that dictates the dynamics of various fluid flows, encompassing turbulent phenomena as well.
The equation in velocity format can be written as:
\begin{equation}
\frac{\partial \mathbf{u}}{\partial t} + (\mathbf{u} \cdot \nabla) \mathbf{u} = -\frac{1}{\rho} \nabla p + \nu \nabla^2 \mathbf{u} + \mathbf{f}, \quad x \in (0,1)^2, t \in (0,T],
\end{equation}
where $\mathbf{f}$ is the \text{external force}, $\nu$ represents kinematic viscosity, $p$ is the pressure term and $\mathbf{u}$ is the velocity vector. 
The problem studied in this work follows the previous work of \citet{li2020fourier}, where the target is to predict the vorticity: $\omega=\partial u_y / \partial x - \partial u_x / \partial y$ given a fixed time horizon $T$ and the initial value $\omega_0$ sampled from a Gaussian random field. The dataset is generated on a 256 $\times$ 256 grid and sub-sampled to 64 $\times$ 64 for training and testing. 

Generally, when the viscosity coefficient $\nu$ is lower, the dynamics become more chaotic, posing a greater challenge for learning. The results for the Navier-Stokes experiments are presented in Table \ref{table: ns}.
In the case of Navier-Stokes, the model is trained for 125,000 iterations with a batch size of 4. For the size of each Navier-Stokes dataset, NS2-full contains 9800/200 (train/test) samples; each of the rest datasets contains 1000/200 samples. Solving a complex equation like Navier-Stokes, the Hyena neural operator significantly outperforms the Fourier neural operator when tested on different viscosities $\nu = 10^{-3},10^{-4},10^{-5}$ with varying $T$ on both large dataset and small dataset . For viscosity such as $\nu=10^{-5}$, where the flow change is more complicated compared to other viscosities, the Hyena operator can keep up with temporal changes due to its ability to capture the global interaction with long convoluions. By applying the curriculum strategy to train the time-dependent data, the model was able to learn the solution more efficiently and converge slightly faster.

\begin{table}[t]
\vspace{-1mm}
\begin{center}
\scalebox{0.95}{
\begin{tabular}{cc!{\vrule width \lightrulewidth}ccc!{\vrule width \lightrulewidth}c} 
\toprule
\multicolumn{2}{c!{\vrule width \lightrulewidth}}{Data settings}          & \multicolumn{4}{c}{Relative $L_2$ norm}                                          \\ 
\midrule
    Case       & $\nu, T$  & FNO-2D  & FNO-3D & U-Net & HNO  \\ 
\midrule
NS1 & $1\times 10^{-3}, 50$ &   0.0128     &    0.0086    &   0.0245                        &         \textbf{0.0069}          \\

NS2-part & $1\times 10^{-4}, 30$ &    0.1559    &    0.1918    &      0.2051      &       \textbf{0.1245}         \\
NS2-full & $1\times 10^{-4}, 30$   &    0.0834    &       0.0820 &   0.1190                    & \textbf{0.0681}                \\
NS3    & $1\times 10^{-5}, 20$  &    0.1556    &    0.1893    & 0.1982        &                \textbf{0.1415}              \\
\midrule
\multicolumn{2}{c!{\vrule width \lightrulewidth}}{\# of parameters (M)}   & 0.41  & 6.56   & 24.95                        & 9.23                 \\
\bottomrule
\end{tabular}}
\caption{ Relative $L_{2}$ norm for Navier-Stokes equation benchmark with a fixed resolution of 64x64. \textbf{Bold} indicates best performance. }\label{table: ns}
\vspace{+2mm}
\end{center}
\vspace{-2mm}
\end{table}
\section{Conclusion} 
In this study, we present the Hyena neural operator, a subquadratic state-space model for learning the solution of PDEs. The data-controlled linear operator demonstrated promising performance and achieved competitive outcomes when compared to alternative approaches. Future work for HNO includes downsampling the high-resolution data in latent space by using contracting-expanding architecture such as Unet\cite{ronneberger2015u}.
Other directions include using tokenized equations to learn physically relevant information\cite{lorsung2023physics} and improve the HNO further.

    
    
\begin{acknowledgement}

This work is supported by the National Science Foundation under Grant No. 1953222.

\end{acknowledgement}

\bibliography{reference}

\providecommand{\latin}[1]{#1}
\makeatletter
\providecommand{\doi}
  {\begingroup\let\do\@makeother\dospecials
  \catcode`\{=1 \catcode`\}=2 \doi@aux}
\providecommand{\doi@aux}[1]{\endgroup\texttt{#1}}
\makeatother
\providecommand*\mcitethebibliography{\thebibliography}
\csname @ifundefined\endcsname{endmcitethebibliography}
  {\let\endmcitethebibliography\endthebibliography}{}
\begin{mcitethebibliography}{51}
\providecommand*\natexlab[1]{#1}
\providecommand*\mciteSetBstSublistMode[1]{}
\providecommand*\mciteSetBstMaxWidthForm[2]{}
\providecommand*\mciteBstWouldAddEndPuncttrue
  {\def\EndOfBibitem{\unskip.}}
\providecommand*\mciteBstWouldAddEndPunctfalse
  {\let\EndOfBibitem\relax}
\providecommand*\mciteSetBstMidEndSepPunct[3]{}
\providecommand*\mciteSetBstSublistLabelBeginEnd[3]{}
\providecommand*\EndOfBibitem{}
\mciteSetBstSublistMode{f}
\mciteSetBstMaxWidthForm{subitem}{(\alph{mcitesubitemcount})}
\mciteSetBstSublistLabelBeginEnd
  {\mcitemaxwidthsubitemform\space}
  {\relax}
  {\relax}

\bibitem[Foucart \latin{et~al.}(2022)Foucart, Charous, and
  Lermusiaux]{foucart2022deep}
Foucart,~C.; Charous,~A.; Lermusiaux,~P.~F. Deep Reinforcement Learning for
  Adaptive Mesh Refinement. \emph{arXiv preprint arXiv:2209.12351}
  \textbf{2022}, \relax
\mciteBstWouldAddEndPunctfalse
\mciteSetBstMidEndSepPunct{\mcitedefaultmidpunct}
{}{\mcitedefaultseppunct}\relax
\EndOfBibitem
\bibitem[Yang \latin{et~al.}(2023)Yang, Dzanic, Petersen, Kudo, Mittal, Tomov,
  Camier, Zhao, Zha, Kolev, Anderson, and Faissol]{pmlr-v206-yang23e}
Yang,~J.; Dzanic,~T.; Petersen,~B.; Kudo,~J.; Mittal,~K.; Tomov,~V.;
  Camier,~J.-S.; Zhao,~T.; Zha,~H.; Kolev,~T.; Anderson,~R.; Faissol,~D.
  Reinforcement Learning for Adaptive Mesh Refinement. Proceedings of The 26th
  International Conference on Artificial Intelligence and Statistics. 2023; pp
  5997--6014\relax
\mciteBstWouldAddEndPuncttrue
\mciteSetBstMidEndSepPunct{\mcitedefaultmidpunct}
{\mcitedefaultendpunct}{\mcitedefaultseppunct}\relax
\EndOfBibitem
\bibitem[Lorsung and Barati~Farimani(2023)Lorsung, and
  Barati~Farimani]{meshdqn}
Lorsung,~C.; Barati~Farimani,~A. {Mesh deep Q network: A deep reinforcement
  learning framework for improving meshes in computational fluid dynamics}.
  \emph{AIP Advances} \textbf{2023}, \emph{13}, 015026\relax
\mciteBstWouldAddEndPuncttrue
\mciteSetBstMidEndSepPunct{\mcitedefaultmidpunct}
{\mcitedefaultendpunct}{\mcitedefaultseppunct}\relax
\EndOfBibitem
\bibitem[Pant \latin{et~al.}(2021)Pant, Doshi, Bahl, and
  Barati~Farimani]{dl_rom}
Pant,~P.; Doshi,~R.; Bahl,~P.; Barati~Farimani,~A. {Deep learning for reduced
  order modelling and efficient temporal evolution of fluid simulations}.
  \emph{Physics of Fluids} \textbf{2021}, \emph{33}, 107101\relax
\mciteBstWouldAddEndPuncttrue
\mciteSetBstMidEndSepPunct{\mcitedefaultmidpunct}
{\mcitedefaultendpunct}{\mcitedefaultseppunct}\relax
\EndOfBibitem
\bibitem[Hemmasian \latin{et~al.}(2023)Hemmasian, Ogoke, Akbari, Malen, Beuth,
  and Farimani]{hemmasian_surrogate_2023}
Hemmasian,~A.; Ogoke,~F.; Akbari,~P.; Malen,~J.; Beuth,~J.; Farimani,~A.~B.
  Surrogate modeling of melt pool temperature field using deep learning.
  \emph{Additive Manufacturing Letters} \textbf{2023}, \emph{5}, 100123\relax
\mciteBstWouldAddEndPuncttrue
\mciteSetBstMidEndSepPunct{\mcitedefaultmidpunct}
{\mcitedefaultendpunct}{\mcitedefaultseppunct}\relax
\EndOfBibitem
\bibitem[Farimani \latin{et~al.}(2017)Farimani, Gomes, and
  Pande]{farimani2017deep}
Farimani,~A.~B.; Gomes,~J.; Pande,~V.~S. Deep Learning the Physics of Transport
  Phenomena. 2017\relax
\mciteBstWouldAddEndPuncttrue
\mciteSetBstMidEndSepPunct{\mcitedefaultmidpunct}
{\mcitedefaultendpunct}{\mcitedefaultseppunct}\relax
\EndOfBibitem
\bibitem[Gao \latin{et~al.}(2021)Gao, Sun, and Wang]{fluid_superresolution}
Gao,~H.; Sun,~L.; Wang,~J.-X. {Super-resolution and denoising of fluid flow
  using physics-informed convolutional neural networks without high-resolution
  labels}. \emph{Physics of Fluids} \textbf{2021}, \emph{33}, 073603\relax
\mciteBstWouldAddEndPuncttrue
\mciteSetBstMidEndSepPunct{\mcitedefaultmidpunct}
{\mcitedefaultendpunct}{\mcitedefaultseppunct}\relax
\EndOfBibitem
\bibitem[Xie \latin{et~al.}(2018)Xie, Franz, Chu, and Thuerey]{volumetric_gan}
Xie,~Y.; Franz,~E.; Chu,~M.; Thuerey,~N. TempoGAN: A Temporally Coherent,
  Volumetric GAN for Super-Resolution Fluid Flow. \emph{ACM Trans. Graph.}
  \textbf{2018}, \emph{37}\relax
\mciteBstWouldAddEndPuncttrue
\mciteSetBstMidEndSepPunct{\mcitedefaultmidpunct}
{\mcitedefaultendpunct}{\mcitedefaultseppunct}\relax
\EndOfBibitem
\bibitem[Werhahn \latin{et~al.}(2019)Werhahn, Xie, Chu, and
  Thuerey]{Werhahn_2019}
Werhahn,~M.; Xie,~Y.; Chu,~M.; Thuerey,~N. A Multi-Pass {GAN} for Fluid Flow
  Super-Resolution. \emph{Proceedings of the {ACM} on Computer Graphics and
  Interactive Techniques} \textbf{2019}, \emph{2}, 1--21\relax
\mciteBstWouldAddEndPuncttrue
\mciteSetBstMidEndSepPunct{\mcitedefaultmidpunct}
{\mcitedefaultendpunct}{\mcitedefaultseppunct}\relax
\EndOfBibitem
\bibitem[Shu \latin{et~al.}(2023)Shu, Li, and
  Farimani]{shu_physics-informed_2023}
Shu,~D.; Li,~Z.; Farimani,~A.~B. A physics-informed diffusion model for
  high-fidelity flow field reconstruction. \emph{Journal of Computational
  Physics} \textbf{2023}, \emph{478}, 111972\relax
\mciteBstWouldAddEndPuncttrue
\mciteSetBstMidEndSepPunct{\mcitedefaultmidpunct}
{\mcitedefaultendpunct}{\mcitedefaultseppunct}\relax
\EndOfBibitem
\bibitem[Yang and Sommer(2023)Yang, and Sommer]{yang2023denoising}
Yang,~G.; Sommer,~S. A Denoising Diffusion Model for Fluid Field Prediction.
  2023\relax
\mciteBstWouldAddEndPuncttrue
\mciteSetBstMidEndSepPunct{\mcitedefaultmidpunct}
{\mcitedefaultendpunct}{\mcitedefaultseppunct}\relax
\EndOfBibitem
\bibitem[Wang \latin{et~al.}(2023)Wang, Plechac, and Knap]{wang2023generative}
Wang,~T.; Plechac,~P.; Knap,~J. Generative diffusion learning for parametric
  partial differential equations. 2023\relax
\mciteBstWouldAddEndPuncttrue
\mciteSetBstMidEndSepPunct{\mcitedefaultmidpunct}
{\mcitedefaultendpunct}{\mcitedefaultseppunct}\relax
\EndOfBibitem
\bibitem[Jadhav \latin{et~al.}()Jadhav, Berthel, Hu, Panat, Beuth, and
  Barati~Farimani]{jadhavstressd}
Jadhav,~Y.; Berthel,~J.; Hu,~C.; Panat,~R.; Beuth,~J.; Barati~Farimani,~A.
  Stressd: 2d Stress Estimation Using Denoising Diffusion Model.
  \emph{Available at SSRN 4478596} \relax
\mciteBstWouldAddEndPunctfalse
\mciteSetBstMidEndSepPunct{\mcitedefaultmidpunct}
{}{\mcitedefaultseppunct}\relax
\EndOfBibitem
\bibitem[Lu \latin{et~al.}(2019)Lu, Jin, and Karniadakis]{lu2019deeponet}
Lu,~L.; Jin,~P.; Karniadakis,~G.~E. Deeponet: Learning nonlinear operators for
  identifying differential equations based on the universal approximation
  theorem of operators. \emph{arXiv preprint arXiv:1910.03193} \textbf{2019},
  \relax
\mciteBstWouldAddEndPunctfalse
\mciteSetBstMidEndSepPunct{\mcitedefaultmidpunct}
{}{\mcitedefaultseppunct}\relax
\EndOfBibitem
\bibitem[Wang \latin{et~al.}(2021)Wang, Wang, and Perdikaris]{wang2021learning}
Wang,~S.; Wang,~H.; Perdikaris,~P. Learning the solution operator of parametric
  partial differential equations with physics-informed DeepOnets. 2021\relax
\mciteBstWouldAddEndPuncttrue
\mciteSetBstMidEndSepPunct{\mcitedefaultmidpunct}
{\mcitedefaultendpunct}{\mcitedefaultseppunct}\relax
\EndOfBibitem
\bibitem[Jin \latin{et~al.}(2022)Jin, Meng, and Lu]{jin2022mionet}
Jin,~P.; Meng,~S.; Lu,~L. MIONet: Learning multiple-input operators via tensor
  product. 2022\relax
\mciteBstWouldAddEndPuncttrue
\mciteSetBstMidEndSepPunct{\mcitedefaultmidpunct}
{\mcitedefaultendpunct}{\mcitedefaultseppunct}\relax
\EndOfBibitem
\bibitem[Kovachki \latin{et~al.}(2023)Kovachki, Li, Liu, Azizzadenesheli,
  Bhattacharya, Stuart, and Anandkumar]{kovachki2023neural}
Kovachki,~N.; Li,~Z.; Liu,~B.; Azizzadenesheli,~K.; Bhattacharya,~K.;
  Stuart,~A.; Anandkumar,~A. Neural Operator: Learning Maps Between Function
  Spaces. 2023\relax
\mciteBstWouldAddEndPuncttrue
\mciteSetBstMidEndSepPunct{\mcitedefaultmidpunct}
{\mcitedefaultendpunct}{\mcitedefaultseppunct}\relax
\EndOfBibitem
\bibitem[Li \latin{et~al.}(2020)Li, Kovachki, Azizzadenesheli, Liu,
  Bhattacharya, Stuart, and Anandkumar]{li2020neural}
Li,~Z.; Kovachki,~N.; Azizzadenesheli,~K.; Liu,~B.; Bhattacharya,~K.;
  Stuart,~A.; Anandkumar,~A. Neural Operator: Graph Kernel Network for Partial
  Differential Equations. 2020\relax
\mciteBstWouldAddEndPuncttrue
\mciteSetBstMidEndSepPunct{\mcitedefaultmidpunct}
{\mcitedefaultendpunct}{\mcitedefaultseppunct}\relax
\EndOfBibitem
\bibitem[Li \latin{et~al.}(2020)Li, Kovachki, Azizzadenesheli, Liu,
  Bhattacharya, Stuart, and Anandkumar]{li2020fourier}
Li,~Z.; Kovachki,~N.; Azizzadenesheli,~K.; Liu,~B.; Bhattacharya,~K.;
  Stuart,~A.; Anandkumar,~A. Fourier neural operator for parametric partial
  differential equations. \emph{arXiv preprint arXiv:2010.08895} \textbf{2020},
  \relax
\mciteBstWouldAddEndPunctfalse
\mciteSetBstMidEndSepPunct{\mcitedefaultmidpunct}
{}{\mcitedefaultseppunct}\relax
\EndOfBibitem
\bibitem[Tran \latin{et~al.}(2023)Tran, Mathews, Xie, and
  Ong]{tran2023factorized}
Tran,~A.; Mathews,~A.; Xie,~L.; Ong,~C.~S. Factorized Fourier Neural Operators.
  2023\relax
\mciteBstWouldAddEndPuncttrue
\mciteSetBstMidEndSepPunct{\mcitedefaultmidpunct}
{\mcitedefaultendpunct}{\mcitedefaultseppunct}\relax
\EndOfBibitem
\bibitem[Guibas \latin{et~al.}(2022)Guibas, Mardani, Li, Tao, Anandkumar, and
  Catanzaro]{guibas2022efficient}
Guibas,~J.; Mardani,~M.; Li,~Z.; Tao,~A.; Anandkumar,~A.; Catanzaro,~B.
  Efficient Token Mixing for Transformers via Adaptive Fourier Neural
  Operators. International Conference on Learning Representations. 2022\relax
\mciteBstWouldAddEndPuncttrue
\mciteSetBstMidEndSepPunct{\mcitedefaultmidpunct}
{\mcitedefaultendpunct}{\mcitedefaultseppunct}\relax
\EndOfBibitem
\bibitem[Li \latin{et~al.}(2023)Li, Zheng, Kovachki, Jin, Chen, Liu,
  Azizzadenesheli, and Anandkumar]{li2023physicsinformed}
Li,~Z.; Zheng,~H.; Kovachki,~N.; Jin,~D.; Chen,~H.; Liu,~B.;
  Azizzadenesheli,~K.; Anandkumar,~A. Physics-Informed Neural Operator for
  Learning Partial Differential Equations. 2023\relax
\mciteBstWouldAddEndPuncttrue
\mciteSetBstMidEndSepPunct{\mcitedefaultmidpunct}
{\mcitedefaultendpunct}{\mcitedefaultseppunct}\relax
\EndOfBibitem
\bibitem[Tripura and Chakraborty(2022)Tripura, and
  Chakraborty]{tripura2022wavelet}
Tripura,~T.; Chakraborty,~S. Wavelet neural operator: a neural operator for
  parametric partial differential equations. \emph{arXiv preprint
  arXiv:2205.02191} \textbf{2022}, \relax
\mciteBstWouldAddEndPunctfalse
\mciteSetBstMidEndSepPunct{\mcitedefaultmidpunct}
{}{\mcitedefaultseppunct}\relax
\EndOfBibitem
\bibitem[Gupta \latin{et~al.}(2021)Gupta, Xiao, and
  Bogdan]{gupta2021multiwavelet}
Gupta,~G.; Xiao,~X.; Bogdan,~P. Multiwavelet-based operator learning for
  differential equations. \emph{Advances in neural information processing
  systems} \textbf{2021}, \emph{34}, 24048--24062\relax
\mciteBstWouldAddEndPuncttrue
\mciteSetBstMidEndSepPunct{\mcitedefaultmidpunct}
{\mcitedefaultendpunct}{\mcitedefaultseppunct}\relax
\EndOfBibitem
\bibitem[Cao(2021)]{cao2021choose}
Cao,~S. Choose a transformer: Fourier or galerkin. \emph{Advances in neural
  information processing systems} \textbf{2021}, \emph{34}, 24924--24940\relax
\mciteBstWouldAddEndPuncttrue
\mciteSetBstMidEndSepPunct{\mcitedefaultmidpunct}
{\mcitedefaultendpunct}{\mcitedefaultseppunct}\relax
\EndOfBibitem
\bibitem[Li \latin{et~al.}(2022)Li, Meidani, and Farimani]{li2022transformer}
Li,~Z.; Meidani,~K.; Farimani,~A.~B. Transformer for partial differential
  equations' operator learning. \emph{arXiv preprint arXiv:2205.13671}
  \textbf{2022}, \relax
\mciteBstWouldAddEndPunctfalse
\mciteSetBstMidEndSepPunct{\mcitedefaultmidpunct}
{}{\mcitedefaultseppunct}\relax
\EndOfBibitem
\bibitem[Su \latin{et~al.}(2022)Su, Lu, Pan, Murtadha, Wen, and
  Liu]{su2022roformer}
Su,~J.; Lu,~Y.; Pan,~S.; Murtadha,~A.; Wen,~B.; Liu,~Y. RoFormer: Enhanced
  Transformer with Rotary Position Embedding. 2022\relax
\mciteBstWouldAddEndPuncttrue
\mciteSetBstMidEndSepPunct{\mcitedefaultmidpunct}
{\mcitedefaultendpunct}{\mcitedefaultseppunct}\relax
\EndOfBibitem
\bibitem[Stachenfeld \latin{et~al.}(2022)Stachenfeld, Fielding, Kochkov,
  Cranmer, Pfaff, Godwin, Cui, Ho, Battaglia, and
  Sanchez-Gonzalez]{stachenfeld2022learned}
Stachenfeld,~K.; Fielding,~D.~B.; Kochkov,~D.; Cranmer,~M.; Pfaff,~T.;
  Godwin,~J.; Cui,~C.; Ho,~S.; Battaglia,~P.; Sanchez-Gonzalez,~A. Learned
  Coarse Models for Efficient Turbulence Simulation. 2022\relax
\mciteBstWouldAddEndPuncttrue
\mciteSetBstMidEndSepPunct{\mcitedefaultmidpunct}
{\mcitedefaultendpunct}{\mcitedefaultseppunct}\relax
\EndOfBibitem
\bibitem[He \latin{et~al.}(2016)He, Zhang, Ren, and Sun]{He_2016_CVPR}
He,~K.; Zhang,~X.; Ren,~S.; Sun,~J. Deep Residual Learning for Image
  Recognition. Proceedings of the IEEE Conference on Computer Vision and
  Pattern Recognition (CVPR). 2016\relax
\mciteBstWouldAddEndPuncttrue
\mciteSetBstMidEndSepPunct{\mcitedefaultmidpunct}
{\mcitedefaultendpunct}{\mcitedefaultseppunct}\relax
\EndOfBibitem
\bibitem[Chen(1984)]{chen1984linear}
Chen,~C.-T. \emph{Linear system theory and design}; Saunders college
  publishing, 1984\relax
\mciteBstWouldAddEndPuncttrue
\mciteSetBstMidEndSepPunct{\mcitedefaultmidpunct}
{\mcitedefaultendpunct}{\mcitedefaultseppunct}\relax
\EndOfBibitem
\bibitem[Gu \latin{et~al.}(2022)Gu, Goel, and Ré]{gu2022efficiently}
Gu,~A.; Goel,~K.; Ré,~C. Efficiently Modeling Long Sequences with Structured
  State Spaces. 2022\relax
\mciteBstWouldAddEndPuncttrue
\mciteSetBstMidEndSepPunct{\mcitedefaultmidpunct}
{\mcitedefaultendpunct}{\mcitedefaultseppunct}\relax
\EndOfBibitem
\bibitem[Gu \latin{et~al.}(2022)Gu, Gupta, Goel, and
  Ré]{gu2022parameterization}
Gu,~A.; Gupta,~A.; Goel,~K.; Ré,~C. On the Parameterization and Initialization
  of Diagonal State Space Models. 2022\relax
\mciteBstWouldAddEndPuncttrue
\mciteSetBstMidEndSepPunct{\mcitedefaultmidpunct}
{\mcitedefaultendpunct}{\mcitedefaultseppunct}\relax
\EndOfBibitem
\bibitem[Gu \latin{et~al.}(2022)Gu, Johnson, Timalsina, Rudra, and
  Ré]{gu2022train}
Gu,~A.; Johnson,~I.; Timalsina,~A.; Rudra,~A.; Ré,~C. How to Train Your HiPPO:
  State Space Models with Generalized Orthogonal Basis Projections. 2022\relax
\mciteBstWouldAddEndPuncttrue
\mciteSetBstMidEndSepPunct{\mcitedefaultmidpunct}
{\mcitedefaultendpunct}{\mcitedefaultseppunct}\relax
\EndOfBibitem
\bibitem[Mehta \latin{et~al.}(2022)Mehta, Gupta, Cutkosky, and
  Neyshabur]{mehta2022long}
Mehta,~H.; Gupta,~A.; Cutkosky,~A.; Neyshabur,~B. Long range language modeling
  via gated state spaces. \emph{arXiv preprint arXiv:2206.13947} \textbf{2022},
  \relax
\mciteBstWouldAddEndPunctfalse
\mciteSetBstMidEndSepPunct{\mcitedefaultmidpunct}
{}{\mcitedefaultseppunct}\relax
\EndOfBibitem
\bibitem[Dao \latin{et~al.}(2022)Dao, Fu, Saab, Thomas, Rudra, and
  R{\'e}]{dao2022hungry}
Dao,~T.; Fu,~D.~Y.; Saab,~K.~K.; Thomas,~A.~W.; Rudra,~A.; R{\'e},~C. Hungry
  Hungry Hippos: Towards Language Modeling with State Space Models. \emph{arXiv
  preprint arXiv:2212.14052} \textbf{2022}, \relax
\mciteBstWouldAddEndPunctfalse
\mciteSetBstMidEndSepPunct{\mcitedefaultmidpunct}
{}{\mcitedefaultseppunct}\relax
\EndOfBibitem
\bibitem[Poli \latin{et~al.}(2023)Poli, Massaroli, Nguyen, Fu, Dao, Baccus,
  Bengio, Ermon, and Ré]{poli2023hyena}
Poli,~M.; Massaroli,~S.; Nguyen,~E.; Fu,~D.~Y.; Dao,~T.; Baccus,~S.;
  Bengio,~Y.; Ermon,~S.; Ré,~C. Hyena Hierarchy: Towards Larger Convolutional
  Language Models. 2023\relax
\mciteBstWouldAddEndPuncttrue
\mciteSetBstMidEndSepPunct{\mcitedefaultmidpunct}
{\mcitedefaultendpunct}{\mcitedefaultseppunct}\relax
\EndOfBibitem
\bibitem[Tay \latin{et~al.}(2020)Tay, Dehghani, Abnar, Shen, Bahri, Pham, Rao,
  Yang, Ruder, and Metzler]{tay2020long}
Tay,~Y.; Dehghani,~M.; Abnar,~S.; Shen,~Y.; Bahri,~D.; Pham,~P.; Rao,~J.;
  Yang,~L.; Ruder,~S.; Metzler,~D. Long range arena: A benchmark for efficient
  transformers. \emph{arXiv preprint arXiv:2011.04006} \textbf{2020}, \relax
\mciteBstWouldAddEndPunctfalse
\mciteSetBstMidEndSepPunct{\mcitedefaultmidpunct}
{}{\mcitedefaultseppunct}\relax
\EndOfBibitem
\bibitem[Mildenhall \latin{et~al.}(2021)Mildenhall, Srinivasan, Tancik, Barron,
  Ramamoorthi, and Ng]{mildenhall2021nerf}
Mildenhall,~B.; Srinivasan,~P.~P.; Tancik,~M.; Barron,~J.~T.; Ramamoorthi,~R.;
  Ng,~R. Nerf: Representing scenes as neural radiance fields for view
  synthesis. \emph{Communications of the ACM} \textbf{2021}, \emph{65},
  99--106\relax
\mciteBstWouldAddEndPuncttrue
\mciteSetBstMidEndSepPunct{\mcitedefaultmidpunct}
{\mcitedefaultendpunct}{\mcitedefaultseppunct}\relax
\EndOfBibitem
\bibitem[Sitzmann \latin{et~al.}(2020)Sitzmann, Martel, Bergman, Lindell, and
  Wetzstein]{sitzmann2020implicit}
Sitzmann,~V.; Martel,~J. N.~P.; Bergman,~A.~W.; Lindell,~D.~B.; Wetzstein,~G.
  Implicit Neural Representations with Periodic Activation Functions.
  2020\relax
\mciteBstWouldAddEndPuncttrue
\mciteSetBstMidEndSepPunct{\mcitedefaultmidpunct}
{\mcitedefaultendpunct}{\mcitedefaultseppunct}\relax
\EndOfBibitem
\bibitem[Romero \latin{et~al.}(2022)Romero, Kuzina, Bekkers, Tomczak, and
  Hoogendoorn]{romero2022ckconv}
Romero,~D.~W.; Kuzina,~A.; Bekkers,~E.~J.; Tomczak,~J.~M.; Hoogendoorn,~M.
  CKConv: Continuous Kernel Convolution For Sequential Data. 2022\relax
\mciteBstWouldAddEndPuncttrue
\mciteSetBstMidEndSepPunct{\mcitedefaultmidpunct}
{\mcitedefaultendpunct}{\mcitedefaultseppunct}\relax
\EndOfBibitem
\bibitem[Tancik \latin{et~al.}(2020)Tancik, Srinivasan, Mildenhall,
  Fridovich-Keil, Raghavan, Singhal, Ramamoorthi, Barron, and
  Ng]{tancik2020fourier}
Tancik,~M.; Srinivasan,~P.; Mildenhall,~B.; Fridovich-Keil,~S.; Raghavan,~N.;
  Singhal,~U.; Ramamoorthi,~R.; Barron,~J.; Ng,~R. Fourier features let
  networks learn high frequency functions in low dimensional domains.
  \emph{Advances in Neural Information Processing Systems} \textbf{2020},
  \emph{33}, 7537--7547\relax
\mciteBstWouldAddEndPuncttrue
\mciteSetBstMidEndSepPunct{\mcitedefaultmidpunct}
{\mcitedefaultendpunct}{\mcitedefaultseppunct}\relax
\EndOfBibitem
\bibitem[Ba \latin{et~al.}(2016)Ba, Kiros, and Hinton]{ba2016layer}
Ba,~J.~L.; Kiros,~J.~R.; Hinton,~G.~E. Layer Normalization. 2016\relax
\mciteBstWouldAddEndPuncttrue
\mciteSetBstMidEndSepPunct{\mcitedefaultmidpunct}
{\mcitedefaultendpunct}{\mcitedefaultseppunct}\relax
\EndOfBibitem
\bibitem[Rahimi and Recht(2007)Rahimi, and Recht]{random_fourier_feature}
Rahimi,~A.; Recht,~B. Random Features for Large-Scale Kernel Machines. Advances
  in Neural Information Processing Systems. 2007\relax
\mciteBstWouldAddEndPuncttrue
\mciteSetBstMidEndSepPunct{\mcitedefaultmidpunct}
{\mcitedefaultendpunct}{\mcitedefaultseppunct}\relax
\EndOfBibitem
\bibitem[Kingma and Ba(2014)Kingma, and Ba]{kingma2014adam}
Kingma,~D.~P.; Ba,~J. Adam: A method for stochastic optimization. \emph{arXiv
  preprint arXiv:1412.6980} \textbf{2014}, \relax
\mciteBstWouldAddEndPunctfalse
\mciteSetBstMidEndSepPunct{\mcitedefaultmidpunct}
{}{\mcitedefaultseppunct}\relax
\EndOfBibitem
\bibitem[Loshchilov and Hutter(2017)Loshchilov, and Hutter]{loshchilov2017sgdr}
Loshchilov,~I.; Hutter,~F. SGDR: Stochastic Gradient Descent with Warm
  Restarts. 2017\relax
\mciteBstWouldAddEndPuncttrue
\mciteSetBstMidEndSepPunct{\mcitedefaultmidpunct}
{\mcitedefaultendpunct}{\mcitedefaultseppunct}\relax
\EndOfBibitem
\bibitem[Hendrycks and Gimpel(2020)Hendrycks, and
  Gimpel]{hendrycks2020gaussian}
Hendrycks,~D.; Gimpel,~K. Gaussian Error Linear Units (GELUs). 2020\relax
\mciteBstWouldAddEndPuncttrue
\mciteSetBstMidEndSepPunct{\mcitedefaultmidpunct}
{\mcitedefaultendpunct}{\mcitedefaultseppunct}\relax
\EndOfBibitem
\bibitem[Takamoto \latin{et~al.}(2022)Takamoto, Praditia, Leiteritz, MacKinlay,
  Alesiani, Pfl{\"u}ger, and Niepert]{takamoto2022pdebench}
Takamoto,~M.; Praditia,~T.; Leiteritz,~R.; MacKinlay,~D.; Alesiani,~F.;
  Pfl{\"u}ger,~D.; Niepert,~M. PDEBench: An extensive benchmark for scientific
  machine learning. \emph{Advances in Neural Information Processing Systems}
  \textbf{2022}, \emph{35}, 1596--1611\relax
\mciteBstWouldAddEndPuncttrue
\mciteSetBstMidEndSepPunct{\mcitedefaultmidpunct}
{\mcitedefaultendpunct}{\mcitedefaultseppunct}\relax
\EndOfBibitem
\bibitem[Krishnapriyan \latin{et~al.}(2021)Krishnapriyan, Gholami, Zhe, Kirby,
  and Mahoney]{krishnapriyan2021characterizing}
Krishnapriyan,~A.; Gholami,~A.; Zhe,~S.; Kirby,~R.; Mahoney,~M.~W.
  Characterizing possible failure modes in physics-informed neural networks.
  \emph{Advances in Neural Information Processing Systems} \textbf{2021},
  \emph{34}, 26548--26560\relax
\mciteBstWouldAddEndPuncttrue
\mciteSetBstMidEndSepPunct{\mcitedefaultmidpunct}
{\mcitedefaultendpunct}{\mcitedefaultseppunct}\relax
\EndOfBibitem
\bibitem[Ronneberger \latin{et~al.}(2015)Ronneberger, Fischer, and
  Brox]{ronneberger2015u}
Ronneberger,~O.; Fischer,~P.; Brox,~T. U-net: Convolutional networks for
  biomedical image segmentation. Medical Image Computing and Computer-Assisted
  Intervention--MICCAI 2015: 18th International Conference, Munich, Germany,
  October 5-9, 2015, Proceedings, Part III 18. 2015; pp 234--241\relax
\mciteBstWouldAddEndPuncttrue
\mciteSetBstMidEndSepPunct{\mcitedefaultmidpunct}
{\mcitedefaultendpunct}{\mcitedefaultseppunct}\relax
\EndOfBibitem
\bibitem[Lorsung \latin{et~al.}(2023)Lorsung, Li, and
  Farimani]{lorsung2023physics}
Lorsung,~C.; Li,~Z.; Farimani,~A.~B. Physics Informed Token Transformer.
  \emph{arXiv preprint arXiv:2305.08757} \textbf{2023}, \relax
\mciteBstWouldAddEndPunctfalse
\mciteSetBstMidEndSepPunct{\mcitedefaultmidpunct}
{}{\mcitedefaultseppunct}\relax
\EndOfBibitem
\end{mcitethebibliography}
\newpage
\appendix
\section*{Appendix A}

\section{Model implementation details}
\label{appendix sec:1}
Below we provide the implementation details of models used in Diffusion-Reaction and Navier-Stokes problems. All the models are implemented in PyTorch 


\subsection{Hyperparameters}

\begin{table}[h]
\centering
\begin{tabular}{c!{\vrule width \lightrulewidth}ccccc} 
\toprule
Problem & \begin{tabular}[c]{@{}c@{}}Input encoder \\top\end{tabular} & Hyena hidden dim & Hyena FFN~~ & \begin{tabular}[c]{@{}c@{}}Input encoder \\bottom\end{tabular} \\ 
\midrule
1D Diffusion-Reaction & {[}2, 128] & 128$\times$8 & {[}128, 256, 128]  & {[}128,~ 128] \\
2D Navier-Stokes & {[}12, 96] & 96$\times$8 & {[}96, 192, 96] & {[}96, 192] \\
\bottomrule
\end{tabular}
\caption{Architecture of input encoder. SA denotes self-attention, FFN denotes feed forward network.}
\end{table}
\begin{table}
\scalebox{0.90}{
\begin{tabular}{c|ccccc} 
\toprule
Problem              & \begin{tabular}[c]{@{}c@{}}Query encoder \\top\end{tabular} & CA hidden dim & \begin{tabular}[c]{@{}c@{}}Hyena Hidden \\dim\end{tabular} & CA FFN & \begin{tabular}[c]{@{}c@{}}Query encoder \\bottom\end{tabular}  \\ 
\midrule
1D Diffusion-Reaction & {[}1, 128, 128]                                               & 128$\times$8          &128$\times$3 & {[}128, 256, 128]   & -                                                               \\
2D Navier-Stokes     & {[}2, 192, 192]                       & 192$\times$4         &192$\times$3  & {[}192, 384, 196] & {[}192, 384]                                                    \\
\bottomrule
\end{tabular}}
\vspace{+1mm}
\caption{Architecture of query encoder. CA denotes cross-attention.}
\end{table}
\begin{table}
\scalebox{0.9}{
\begin{tabular}{c|cccc} 
\toprule
Problem              & Propagator             & Decoder             & Hyena hidden dim & Total \# of params (M)  \\ 
\midrule
1D Diffusion-Reaction & {[}128, 128, 128]$\times$3   & {[}128,~ 64, 1]      & 128$\times$8 & 5.61                    \\
2D Navier-Stokes     & {[}384, 384, 384] & {[}384, 192, 96, 1] & 192$\times$8 & 9.22                    \\
\bottomrule
\end{tabular}}
\caption{ Architecture of propagator and decoder. The number of parameters is the total number including input/query encoders. For 1D Diffusion-Reaction' equation we use 3 unshared MLPs for propagating the dynamics.}
\end{table}
\end{document}